# Stabilization and Trajectory Control of a Quadrotor with Uncertain Suspended Load

**Xu Zhou**
Department of Mechanical Engineering
Colorado School of Mines
Golden, CO, 80401
e-mail: xuzhou@mines.edu

**Xiaoli Zhang[1]**
Department of Mechanical Engineering
Colorado School of Mines
Golden, CO, 80401
e-mail: xlzhang@mines.edu
ASME Membership

**Jiucai Zhang**
National Renewable Energy Laboratory
Golden, CO, 80401
e-mail: Jiucai.Zhang@nrel.gov

**Rui Liu**
Department of Mechanical Engineering
Colorado School of Mines
Golden, CO, 80401
e-mail: rliu@mines.edu

**ABSTRACT**

*Stabilization and trajectory control of a quadrotor carrying a suspended load with a fixed known mass has been extensively studied in recent years. However, the load mass is not always known beforehand or may vary during the practical transportations. This mass uncertainty brings uncertain disturbances to the quadrotor system, causing existing controllers to have worse stability and trajectory tracking performance. To improve the quadrotor stability and trajectory tracking capability in this situation, we fully investigate*

---

[1] Corresponding author

*the impacts of the uncertain load mass on the quadrotor. By comparing the performances of three different controllers -- the proportional-derivative (PD) controller, the sliding mode controller (SMC), and the model predictive controller (MPC) -- stabilization rather than trajectory tracking error is proved to be the main influence in the load mass uncertainty. A critical motion mass exists for the quadrotor to maintain a desired transportation performance. Moreover, simulation results verify that a controller with strong robustness against disturbances is a good choice for practical applications.*

## 1. INTRODUCTION

Aerial load transportation has recently received considerable attentions as an important application of the physical interaction between unmanned aerial vehicles (UAVs) and the surrounding environment [1, 2]. Specifically, quadrotor UAVs can play an important role in delivering loads to some dangerous or hardly inaccessible places due to their simple mechanical structures as well as desirable hovering and vertical take-off and landing (VTOL) capabilities [3-7]. Even though the quadrotor load transportation is promising, it is still very challenging in real situations. One reason is the quadrotor itself is an underactuated system. With only four inputs with six degrees of freedom (DOF) to control, the quadrotor is easily unstable and inherently difficult to control. As safety is a critical factor in the load transportation, the unstable quadrotor system could behave inappropriately or even crash. Another difficulty is suspended loads can change the quadrotor's own dynamics, thereby bringing external disturbances into the quadrotor model. This could potentially cause existing model-based controllers to have decreased performance or even cause collapse in the trajectory tracking which is another important criterion for load transportation. These open challenges limit stability and trajectory tracking performance of aerial load transportation and therefore hinder its adaptation in practical applications.

Certain load transportation with a prior known load mass for quadrotors has been investigated by many researchers in recent years. These works mainly focus on the stability and trajectory tracking problems. For example, References [3, 4] generated and tracked swing-free trajectories by utilizing the dynamic programming algorithm. A novel coordinate-free form for dynamic motion equations was developed in [5], which was suitable for robot control system design. A geometric nonlinear controller was also presented to asymptotically stabilize the whole system. Reference [8] proved a quadrotor with a cable-suspended load system to be differentially-flat and proposed a nonlinear control method to track the trajectory in a two-dimensional plane. The extension of this method for 3D cases was given in [9]. The change of a quadrotor's center of gravity (CoG) caused by the swing load was dealt with by an adaptive controller in [10]. In [11], more challenging trajectory tracking such as passing through a small window without prior knowledge of the window position was obtained using an iterative Linear-Quadratic-Gaussian algorithm. Reference [12] presented a new motion planning method by generating trajectories with minimal residual oscillations through a finite-sampling, batch reinforcement learning technique. Reference [13] explored the safe and precise operation of the quadrotor with a heavy slung load through a novel nonlinear model predictive controller. In [6, 7, 14-16], the situation in which multiple quadrotors working together to transport a cable suspended load in a three-dimensional space was studied.

However, the practical load mass information could be completely uncertain beforehand or vary during the transportation [17, 18]. From the quadrotor's perspective, this load uncertainty or change results in a correspondingly uncertain or changeable external disturbance. If the controller is not robust against external disturbances, then the stability or trajectory tracking performance could be much worse. Considering the real application needs fast, stable and/or precise load transportation [18], this performance level is unacceptable. To render the quadrotor load transportation practically useful, there is a fundamental need to develop a controller to ensure the stability and trajectory tracking capacity of the quadrotor.

Few works have been reported for quadrotors transporting uncertain loads. Researchers [17] treated the uncertain load as a parametric model uncertainty and developed an adaptive robust controller to compensate for this uncertainty. However, the load in [17] was assumed to be attached to the quadrotor directly, which makes the dynamic model less complex than that of a suspended one. Therefore, the method in [17] will not be considered to apply in the suspended load situation. By treating the uncertain load mass as a nominal mass plus an uncertainty, [18] proposed a fixed-gain nonlinear proportional-derivative (PD) controller for the nominal one and designed another retrospective adaptive controller to compensate for the uncertainty. However, this compensation was considered in only altitude direction. Since the load mass uncertainty affects more than altitude motion, this compensation design may limit its practical applications. More importantly, both [17] and [18] did not fully investigate the vital influence of the uncertain load on the quadrotor.

In this work, our goal is to improve the quadrotor stability and trajectory tracking capability under the influences of uncertain suspended load masses. In order to achieve this goal, we need to substantially understand how the load mass uncertainty impacts the quadrotor. We evaluate the performances of three different controllers -- the PD controller, sliding mode controller (SMC) and model predictive controller (MPC) -- by modifying them to be complete solutions for the quadrotor carrying an uncertain suspended load. After determining the main effects of the uncertain load mass, we further discuss the relationship between the load mass and the quadrotor performance in detail. In short, our main contributions include:

1) Verify that a controller with strong robustness against disturbances is a good choice for the quadrotor carrying an uncertain suspended load. The better performance of SMC and MPC than a conventional PD controller supports this conclusion since SMC and MPC are much more robust against disturbances than PD.

2) Investigate the main influence of the uncertain load on the quadrotor. The conclusion that the uncertain load mainly affects the stabilization dynamic response of the interconnected quadrotor and load system could be used for further practical considerations.

3) Determine the critical motion mass the quadrotor can take while keeping the desired performance. As the load mass increases, the quadrotor is less able to meet the full motion expectations. This critical motion mass could be used for better and more complex controller design.

## 2. MATHEMATICAL MODELING

The controller design should be based on a realistic dynamic model so that the design can be evaluated judicially. Therefore, in this section, we focus on the development of a realistic and comprehensive dynamic model of this interconnected system.

### 2.1 Assumptions

Mathematical models are of great importance for the quadrotor control. Without violating the correctness of the final result, we can make the following assumptions to help us simplify dynamic modeling.

1) The quadrotor body is rigid and symmetrical.
2) The quadrotor's center of gravity is coincided with the body fixed frame origin.
3) The suspension point is exactly the quadrotor's center of gravity.
4) The cable is massless and the cable force is always non-negative and non-negligible.
5) There are no other external disturbances or interactions, such as wind gust, to the quadrotor except for the load.

These assumptions are considered to be sufficient and valid for the realistic representation of the quadrotor with a swing load system, which is used for a nonaggressive trajectory tracking [19]. With assumption (1), we can make the body fixed inertia matrix to be diagonal or, in other words, the product of inertia off the diagonal is zero. Assumption (2) and (3) help to simplify the dynamic modeling by eliminating some unnecessary offsets to the quadrotor's center of gravity. According to assumption (4), the cable should always be taunt to ensure the load is prevented from flying into the upper hemisphere of the quadrotor. Assumption (5) rejects the highly nonlinear aerodynamic affects to ensure the load is the main influence on the quadrotor.

**2.2 Quadrotor Dynamic Modeling**

Since the quadrotor is our main control object, the derivation of its motion equations is necessary and important for further controller design. Considering it is convenient to express the quadrotor's translational motion and rotational motion in the inertial frame and body frame, respectively, we need to choose these two frames first. An inertial frame is used to express the quadrotor's motion in the world coordinate system, so it can be arbitrarily chosen. The body frame selection is based on assumption (2). As shown in Figure 1, Frame {W} is defined as the inertial frame (world frame) and Frame {B} is the body frame.

Based on assumptions (1) - (2), (5) and considering the low-speed fly situation, we can derive the quadrotor dynamics as [20, 21]

$$\dot{p} = v \tag{1}$$

$$m_q \dot{v} = m_q g + \mathbf{R} f \tag{2}$$

$$\dot{\mathbf{R}} = \mathbf{R} S(\Omega) \tag{3}$$

$$\mathbf{I}\dot{\Omega} = S(\mathbf{I}\Omega)\Omega + T \tag{4}$$

where $p = (x, y, z)^T \in \mathbb{R}^3$ is the quadrotor position measured in frame {W}, $v = \dot{p} \in \mathbb{R}^3$ is the linear translational velocity expressed in frame {W}. $\dot{v} \in \mathbb{R}^3$ is the linear translational acceleration expressed in frame {W}. $m_q$ is the quadrotor mass, $g = (0,0,g)^T$ is the acceleration of gravity. $\mathbf{R} \in SO^3$ is a rotation matrix from the body frame {B} to the inertial frame {W}. $f$ is the upwards thrust directed along the negative body-aligned z-axis. $\boldsymbol{\Omega} \in \mathbb{R}^3$ is the angular velocity expressed in frame {B}. $\dot{\boldsymbol{\Omega}} \in \mathbb{R}^3$ is the angular acceleration expressed in frame {B}. $\mathbf{S}(\cdot)$ is the skew-symmetric operator such that $p \times q = \mathbf{S}(p)q$. $\mathbf{I} = diag\{I_x, I_y, I_z\}$ is the body-fixed inertia matrix. $\boldsymbol{T}$ is the applied moments on quadrotor. Eqs. (1) and (2) describe the linear translational motion in the inertial frame. After obtaining the thrust vector in the inertial frame by using rotation matrix R to map it from the body frame to the inertial frame, the resultant force of this thrust vector and the gravity equals to the multiplication of the quadrotor mass and translational acceleration. While it is convenient to use the inertial frame to express linear equations, the body frame is useful in having rotational motion equations. Eqs. (3) and (4) express how the angular velocity and external torques contribute to the angular acceleration.

Extending Eqs. (1) - (4), we can obtain the complete quadrotor dynamic model in Eq. (5) [22, 23]. The three translational accelerations are related to the total thrust generated by four propellers as well as the attitudes. However, the rotational accelerations are only related to the torques generated by propellers and the attitude angles themselves. They have no relationship with translational variables. In other words, the translational movements are affected by the rotational movements while the rotational movements are relatively independent.

$$\begin{cases} \ddot{x} = \dfrac{(\cos\phi\sin\theta\cos\psi + \sin\phi\sin\psi)}{m_q} U_1 \\ \ddot{y} = \dfrac{(\cos\phi\sin\theta\sin\psi - \sin\phi\cos\psi)}{m_q} U_1 \\ \ddot{z} = \dfrac{\cos\phi\cos\psi}{m_q} U_1 + g \\ \ddot{\phi} = \dfrac{I_y - I_z}{I_x}\dot{\theta}\dot{\psi} + \dfrac{l}{I_x} U_2 \\ \ddot{\theta} = \dfrac{I_z - I_x}{I_y}\dot{\phi}\dot{\psi} + \dfrac{l}{I_y} U_3 \\ \ddot{\psi} = \dfrac{I_x - I_y}{I_z}\dot{\theta}\dot{\phi} + \dfrac{1}{I_z} U_4 \end{cases} \qquad (5)$$

where (x, y, z) are the translational displacements in x, y, and z directions, $(\phi, \theta, \psi)$ are the angles of roll, pitch and yaw, which parameterize the orientation of frame {B} with respect to frame {W}. $l$ is the distance between the rotor's center and the quadrotor's CoG. The dot and double dot symbol mean the corresponding velocity and acceleration. $U_1, U_2, U_3, and\ U_4$ are system inputs, which are the combinations of individual rotor thrusts, where $\boldsymbol{f} = [0, 0, U_1]$ and $\boldsymbol{T} = [U_2, U_3, U_4]$.

$$\begin{cases} U_1 = F_1 + F_2 + F_3 + F_4 \\ U_2 = -F_1 + F_3 \\ U_3 = -F_2 + F_4 \\ U_4 = Q_1 - Q_2 + Q_3 - Q_4 \end{cases} \qquad (6)$$

where $F_i$ represents thrust forces generated by four rotors, respectively, and $Q_i$ represents rotor moments. Both are proportional to the square of the angular speed.

**2.3 Slung Load Modeling**

In order to analyze the load's influence on the original quadrotor dynamics, we need to model the slung load as well. From assumptions (3) and (5), we can consider the slung load as a spherical pendulum fixed at a single point. As shown in Figure 2, the origin of the load coordinate system is exactly at the quadrotor's center of gravity, and the axes are parallel to the quadrotor's axes. The combined light blue circles represent the quadrotor and the dark blue circle represents the load. The light blue line between the quadrotor's COG and load means the cable and its length is L. $r$, $s$ and $\zeta$ indicate the relative positions between the quadrotor and load in x, y, and z directions.

Because the load is not allowed to swing to the upper hemisphere above the quadrotor, we need to make sure $\zeta = \sqrt{L^2 - r^2 - s^2}$ is always nonnegative. For the interconnected system, the Euler-Lagrange method [13, 24] is used to compute the load accelerations $\ddot{r}$ and $\ddot{s}$, which are shown in Eq. (8). From this equation, we can see the load translational accelerations depend not only on the quadrotor's translational variables, but also on its own translational variables. This indicates the complex coupling between the quadrotor and load, which will be analyzed as following.

**2.4 Coupling Dynamics Analysis**

Based on assumption (4), the suspended load has a non-negligible influence on the quadrotor, so there is always a cable force between the quadrotor and load. According to Newton's second law of motion, the cable force equals the mass multiply by the absolute acceleration.

$$F_C = \begin{pmatrix} F_{Cx} \\ F_{Cy} \\ F_{Cz} \end{pmatrix} = -m_L \begin{pmatrix} \ddot{x} + \ddot{r} \\ \ddot{y} + \ddot{s} \\ \ddot{z} + \ddot{\zeta} - g(\zeta/L) \end{pmatrix} \qquad (7)$$

where $m_L$ is the load mass and $L$ is the cable length.

Since the load is suspended at the quadrotor's center of gravity, it only affects the quadrotor's translational motion and the rotational motion remains the same. That means the cable force only causes acceleration on the quadrotor's x, y, and z directions.

Also, we assume the yaw angle is a constant value of zero by an independent controller. Combining Eqs. (1) – (4) and (7), we can obtain the interconnected dynamics shown in Eq. (8).

$$\begin{cases} \ddot{x} = \dfrac{\cos\phi \sin\theta}{m_q + m_L} U_1 - \dfrac{m_L}{m_q + m_L} \ddot{r} \\ \ddot{y} = -\dfrac{\sin\phi}{m_q + m_L} U_1 - \dfrac{m_L}{m_q + m_L} \ddot{s} \\ \ddot{z} = \dfrac{\cos\phi \cos\theta}{m_q + m_L} U_1 + \dfrac{m_L}{m_q + m_L} \dfrac{\ddot{r}r + \dot{r}^2 + \ddot{s}s + \dot{s}^2}{\zeta} \\ \qquad + \dfrac{m_L}{m_q + m_L} \dfrac{(\dot{r}r + \dot{s}s)^2}{\zeta^3} + \dfrac{m_L g \frac{\zeta}{L} + m_q g}{m_q + m_L} \\ \ddot{r} = (\zeta^4 \ddot{x} - r\,\zeta^3 \ddot{z} + rs\zeta^2 \ddot{s} + (rL^2 - rs^2)\dot{r}^2 \\ \qquad + (rL^2 - r^3)\dot{s}^2 + 2\dot{r}\dot{s}r^2 s + rg\,\zeta^3)/((s^2 - L^2)\,\zeta^2) \\ \ddot{s} = (\zeta^4 \ddot{y} - s\,\zeta^3 \ddot{z} + rs\zeta^2 \ddot{r} + (sL^2 - sr^2)\dot{s}^2 \\ \qquad + (sL^2 - s^3)\dot{r}^2 + 2\dot{r}\dot{s}s^2 r + sg\,\zeta^3)/((r^2 - L^2)\,\zeta^2) \end{cases} \qquad (8)$$

Comparing to the original quadrotor dynamics in Eq. (5), we can treat the new r and s related terms in Eq. (8) as external disturbances to the quadrotor. The details will be introduced in Section 3. In this way, we can transform the uncertain suspended load problem back to the quadrotor control problem with external disturbances.

**3. CONTROL ALGORITHM DESIGN**

The goal of the control system design is to make a quadrotor stably and robustly track a desired trajectory under the influences of an uncertain suspended load. In this section, the overall control scheme is firstly explained. Owing to the quadrotor's underactuation characteristic, the overall control scheme includes an outer position controller and an inner attitude controller. The outer position controller generates translational control signal and the inner attitude controller generates the rotational control signals. The position controller also generates desired roll and pitch angles for the attitude controller. This control scheme removes two control targets so that the control outputs equal the inputs, which ensures the interconnected system is fully controllable. Although the overall control scheme is similar for SMC and MPC, their own inherent control designs are different. Section 3.2 and 3.3 introduce SMC and MPC designs, respectively, in detail. The design of SMC is to choose proper sliding surfaces on the basis of tracking error to satisfy necessary sliding conditions and the Lyapunov stability. The MPC design objective is mainly to solve an optimization problem by defining and minimizing an objective function which is related to both the optimal state and control inputs. While SMC is based on the continuous state space equations, MPC needs a discrete-time formation.

**3.1 Overall Control Scheme**

As we described in the previous section, we can treat the uncertain load as a bounded external disturbance to the quadrotor system. Thus, the quadrotor translational dynamic equations can be changed to Eq. (9).

$$\begin{cases} \ddot{x} = \dfrac{\cos\phi \sin\theta}{m_q} U_1 + F_{Cx} \\ \ddot{y} = \dfrac{-\sin\phi}{m_q} U_1 + F_{Cy} \\ \ddot{z} = \dfrac{\cos\phi \cos\theta}{m_q} U_1 + g + F_{Cz} \end{cases} \quad (9)$$

where $F_{Cx}$, $F_{Cy}$ and $F_{Cz}$ are the disturbances caused by the load mass in Eq. (7). We assume these disturbances are bounded because we only consider the safety flight conditions, i.e., the load mass should be inside the range between 0 and the maximum value $M_{max}$. By using Eq. (9), our problem is simplified to a quadrotor control problem with an external disturbance. The external disturbance calculation is based on the load dynamic modeling and coupling dynamics analysis discussed in Section 2. For our situation, we can define the problem as the form

$$\begin{cases} \dot{X} = F(X,t)X + G(X,t)u + F_d \\ Y = H(X,t)X \end{cases} \quad (10)$$

where the state $X = \{x, \dot{x}, y, \dot{y}, z, \dot{z}\}^T$, $Y = \{x, y, z\}^T$, $u = U_1$ and the disturbance $F_d = \{0, F_{Cx}, 0, F_{Cy}, 0, F_{Cz}\}^T$. $F(X,t)$ is the state matrix, $G(X,t)$ is the input matrix, and $H(X,t)$ is the output matrix.

Based on these state space equations, the overall control scheme is shown in Figure 3. It consists of four blocks: the outer position controller, the inner attitude controller, the quadrotor dynamics, and the load dynamics. The outer position controller uses SMC/MPC methods, which will be introduced in the following subsections. It uses the errors between the desired and actual positions as inputs and determines the control signal ($U_1$) for translational motion in quadrotor dynamics as well as the desired roll and pitch angles ($\phi_d$ and $\theta_d$) for the attitude controller to track. Similarly, the SMC/MPC-based inner attitude controller uses the error between the desired and actual angles as inputs and generates three rotational control signals ($U_2, U_3$ and $U_4$) for the quadrotor dynamics block. After obtaining control signals from two controllers, the quadrotor dynamics block calculates the new states defined in Eq. (10) under the disturbance from the load dynamics block. These new states are also used by the load dynamics block and the position controller for another round of control.

**3.2 SMC Algorithm**

SMC is a nonlinear, variable structure control method that alters nonlinear system dynamics by applying a discontinuous control signal that forces the system to "slide" along a cross-section of the system's normal behavior. The basic SMC design procedure is performed in two steps. First, a sliding surface (S) is selected according to the tracking error, while the second step is to make the design of Lyapunov function satisfy the necessary sliding condition [25-27]. The application of SMC in quadrotor dynamics is shown by obtaining the expression for the control input. The definitions of sliding surfaces are

$$\begin{cases} S_\phi = e_2 + \lambda_1 e_1 \\ S_\theta = e_4 + \lambda_2 e_3 \\ S_\psi = e_6 + \lambda_3 e_5 \\ S_x = e_8 + \lambda_4 e_7 \\ S_y = e_{10} + \lambda_5 e_9 \\ S_z = e_{12} + \lambda_6 e_{11} \end{cases} \quad (11)$$

Such that $\lambda_i > 0$, $e_i = x_{id} - x_i$ and $e_{i+1} = \dot{e}_i$, $i \in [1, 3, 5, \ldots, 11]$, where $x_{id}$ is the desired value, $x_i$ is the actual value. Specifically, $x_1 = \phi$, $x_2 = \dot{x}_1 = \dot{\phi}$, $x_3 = \theta$, $x_4 = \dot{x}_3 = \dot{\theta}$, $x_5 = \psi$, $x_6 = \dot{x}_5 = \dot{\psi}$, $x_7 = x$, $x_8 = \dot{x}_7 = \dot{x}$, $x_9 = y$, $x_{10} = \dot{x}_9 = \dot{y}$, $x_{11} = z$, $x_{12} = \dot{x}_{11} = \dot{z}$,

Assuming that $V(S_i) = \frac{1}{2} S_i^2$, then the necessary sliding condition is verified and Lyapunov stability is guaranteed. The chosen law for the attractive surface must satisfy $S_i \dot{S}_i < 0$, so the control law can be obtained below.

$$\begin{cases} U_2 = \dfrac{I_x}{l}\left(-k_1 sign(S_\phi) - \dfrac{I_y - I_z}{I_x} x_4 x_6 + \ddot{\phi}_d + \lambda_1 e_2\right) \\ U_3 = \dfrac{I_y}{l}\left(-k_2 sign(S_\phi) - \dfrac{I_z - I_x}{I_y} x_4 x_6 + \ddot{\theta}_d + \lambda_2 e_4\right) \\ U_4 = I_z\left(-k_3 sign(S_\phi) - \dfrac{I_x - I_y}{I_z} x_4 x_6 + \ddot{\psi}_d + \lambda_3 e_6\right) \\ U_1 = \dfrac{m_q}{\cos x_1 \cos x_3}(-k_4 sign(S_z) - g + \ddot{z}_d + \lambda_4 e_{12}) \\ U_x = \dfrac{m}{U_1}(-k_5 sign(S_x) + \ddot{x}_d + \lambda_5 e_8) \\ U_y = \dfrac{m}{U_1}(-k_6 sign(S_y) + \ddot{y}_d + \lambda_6 e_{10}) \end{cases} \quad (12)$$

where $U_1, U_2, U_3, U_4$ are control signals referred to in Eq. (5). $U_x$ and $U_y$ are control signals for x and y. $U_x$ and $U_y$ are only used for calculating the desired roll and pitch angles, which will be tracked by the inner SMC attitude controller.

After obtaining the control signal $U_1$, we can use Eq. (13) to compute desired roll and pitch angles, which will be used by quadrotor dynamics calculation and the inner SMC attitude controller.

$$\begin{cases} \phi_d = \operatorname{asin}\left(-\dfrac{m_q}{U_1} U_y\right) \\ \theta_d = \operatorname{asin}\left(\dfrac{m_q}{U_1 \cos \phi_d} U_x\right) \end{cases} \quad (13)$$

**3.3 MPC Algorithm**

MPC is an advanced process control method which calculates the future assessment of the system and organizes the control action accordingly. It predicts the future system and control signals in such a manner that it reduces a defined cost function which is an error between the output and desired tracking point over particular prediction horizon [28-30]. The development of MPC can be seen from the block diagram in Figure 4. For our control problem, the plant indicates the quadrotor and the observer refers to Kalman Estimator. Thus, the MPC's main idea is to measure the quadrotor output *y(k)*, estimate an optimal quadrotor state $\hat{x}(k+1)$ by Kalman Estimator, and deliver a new control action to the quadrotor input *u(k)*. The control action is determined by minimizing the cost function related to tracking reference signals and limiting input signals, which will be explained in Eq. (21) in details.

Before implementing the MPC method on the quadrotor, we need to discrete the continuous state space. Similar with SMC method, two control subsystems are used. According to the small angle approximation theory, we can use $\sin\theta \approx \theta$ and $\cos\theta \approx 1$ if $\theta \in [-6°, 6°]$. This approximation helps us obtain the linear discrete-time space equations for translational and rotational subsystems in Eqs. (14) and (15)

$$\begin{bmatrix} x(k+1) \\ v_x(k+1) \\ y(k+1) \\ v_y(k+1) \\ z(k+1) \\ v_z(k+1) \end{bmatrix} = \begin{bmatrix} 1 & \Delta T & 0 & 0 & 0 & 0 \\ 0 & 1 & 0 & 0 & 0 & 0 \\ 0 & 0 & 1 & \Delta T & 0 & 0 \\ 0 & 0 & 0 & 1 & 0 & 0 \\ 0 & 0 & 0 & 0 & 1 & \Delta T \\ 0 & 0 & 0 & 0 & 0 & 1 \end{bmatrix} \begin{bmatrix} x(k) \\ v_x(k) \\ y(k) \\ v_y(k) \\ z(k) \\ v_z(k) \end{bmatrix}$$

$$+ \begin{bmatrix} 0 & 0 & 0 \\ -g\Delta T & 0 & 0 \\ 0 & 0 & 0 \\ 0 & g\Delta T & 0 \\ 0 & 0 & 0 \\ 0 & 0 & \Delta T \end{bmatrix} \begin{bmatrix} \theta(k) \\ \phi(k) \\ G(k) \end{bmatrix}$$

(14)

$$\begin{bmatrix} \phi(k+1) \\ v_\phi(k+1) \\ \theta(k+1) \\ v_\theta(k+1) \\ \psi(k+1) \\ v_\psi(k+1) \end{bmatrix} = \begin{bmatrix} 1 & \Delta T & 0 & 0 & 0 & 0 \\ 0 & 1 & 0 & 0 & 0 & 0 \\ 0 & 0 & 1 & \Delta T & 0 & 0 \\ 0 & 0 & 0 & 1 & 0 & 0 \\ 0 & 0 & 0 & 0 & 1 & \Delta T \\ 0 & 0 & 0 & 0 & 0 & 1 \end{bmatrix} \begin{bmatrix} \phi(k) \\ v_\phi(k) \\ \theta(k) \\ v_\theta(k) \\ \psi(k) \\ v_\psi(k) \end{bmatrix}$$

$$+ \begin{bmatrix} 0 & 0 & 0 \\ \Delta T I_x^{-1} & 0 & 0 \\ 0 & 0 & 0 \\ 0 & \Delta T I_y^{-1} & 0 \\ 0 & 0 & 0 \\ 0 & 0 & \Delta T I_z^{-1} \end{bmatrix} \begin{bmatrix} U_2(k) \\ U_3(k) \\ U_4(k) \end{bmatrix} \quad (15)$$

where $G(k) = g - U_1(k)/m_q$ and $\Delta T$ means the time step for discretization.

These two equations can be regarded as a general discrete state-space system in Eq. (16).

$$\begin{cases} X_{k+1} = AX_k + BU_k + W_k + D_k \\ Y_k = CX_k + V_k \end{cases} \quad (16)$$

where $X_k$ is the state, $Y_k$ is the output, $U_k$ is the input, and $D_k$ is the disturbance vector caused by the load. A is the state matrix, B is the input matrix, C is the output matrix. $W_k$ is the state noise, and $V_k$ is the measurement noise. Both these two noises are assumed to be Gaussian distributed with zero mean and respective covariances of W and V with cross-covariance Z. Taking the translation subsystem in Eq. (14) for example, the state $X_k = [x(k), v_x(k), y(k), v_y(k), z(k), v_z(k)]^T$, the output $Y_k = [x(k), y(k), z(k)]^T$ which makes the state matrix $A = \begin{bmatrix} 1 & \Delta T & 0 & 0 & 0 & 0 \\ 0 & 1 & 0 & 0 & 0 & 0 \\ 0 & 0 & 1 & \Delta T & 0 & 0 \\ 0 & 0 & 0 & 1 & 0 & 0 \\ 0 & 0 & 0 & 0 & 1 & \Delta T \\ 0 & 0 & 0 & 0 & 0 & 1 \end{bmatrix}$, the input matrix $B =$

$$\begin{bmatrix} 0 & 0 & 0 \\ -g\Delta T & 0 & 0 \\ 0 & 0 & 0 \\ 0 & g\Delta T & 0 \\ 0 & 0 & 0 \\ 0 & 0 & \Delta T \end{bmatrix}, \text{ and the output matrix } C = \begin{bmatrix} 1 & 0 & 0 & 0 & 0 & 0 \\ 0 & 0 & 1 & 0 & 0 & 0 \\ 0 & 0 & 0 & 0 & 1 & 0 \end{bmatrix}.$$ The rotational subsystem is similar, so we do not explain it in details here. Given the discrete state-space equations with noises and disturbances, Kalman Estimator is used to make optimal prediction of the next state which is the basis of minimizing the cost function later. Since we do not consider the noise influence in previous SMC and PD designs, which are used for result comparison, we also ignore this part in MPC design. However, W and V in Eq. (16) are required to be positive definite, so we set W and V to be a very small value, 0.0001, and set Z to be 0 [31]. By this way, the noise can be removed to the greatest extent, and the Kalman Estimator mainly deals with the disturbance $D_k$. Let $\hat{X}_{i|j}$ and $\hat{Y}_{i|j}$ represent estimates of the state and output at time i given information up to and including time j where $j \leq i$, then the optimal prediction of next state $\hat{X}_{k+1|k}$ can be obtained as following

$$\begin{cases} \hat{X}_{k+1|k} = A\hat{X}_{k|k-1} + BU_k + K(Y_k - \hat{Y}_{k|k-1}) \\ \hat{Y}_{k|k-1} = C\hat{X}_{k|k-1} \\ K = (APC^T + Z)(CPC^T + V)^{-1} \\ P = W + APA^T - (APC^T + Z)(CPC^T + V)^{-1}(Z^T + CPA^T) \end{cases} \quad (17)$$

The above equation for P is known as the discrete-time-algebraic-Riccati equation, and robust solvers for such equations are available in MATLAB.

After getting the optimal prediction of next state $\hat{X}_{k+1|k}$, we can consider the predictions in the entire prediction horizon. MPC usually requires estimates of the state and/or output over the entire prediction horizon from time *k+1* until *k+N*, and can only make these predictions based on information up to and including the current time $k$.

Optimal estimates from $k+2$ to $k+N$ can be obtained as follows.

$$\begin{cases} \hat{X}_{k+i+1} = A\hat{X}_{k+i|k} + BU_{k+i|k} \\ \hat{Y}_{k+i|k} = C\hat{X}_{k+i|k} \end{cases} \tag{18}$$

where $i = 1,\ldots,N$ and the notation $u_{k+i|k}$ is used to distinguish the actual input at time $k+i$ from that used for prediction purposes. Let $Y_k \triangleq [y_{k+1|k} \quad \cdots \quad y_{k+N|k}]^T$ and $U_k \triangleq [u_{k+1|k} \quad \cdots \quad u_{k+N|k}]^T$, then we can have

$$Y_k = \Lambda \hat{x}_{k+i|k} + \Gamma U_k \tag{19}$$

where

$$\Lambda = \begin{bmatrix} C \\ CA \\ CA^2 \\ \vdots \\ CA^{N-1} \end{bmatrix} \quad \Gamma = \begin{bmatrix} 0 & & & & \\ CB & 0 & & & \\ CAB & CB & 0 & & \\ \vdots & \vdots & \vdots & \ddots & \\ CA^{N-2}B & \cdots & \cdots & \cdots & 0 \end{bmatrix} \tag{20}$$

By defining the trajectory as $r_k$ at time $k$, our control goal is to minimize the objective function in Eq. (21).

$$J(\hat{x}_{k+1|k}, U_k) = \frac{1}{2} \sum_{i=1}^{N} \|\hat{y}_{k+i|k} - r_{k+i}\|_Y^2 + \|u_{k+i|k} - u_{k+i-1|k}\|_S^2 \tag{21}$$

where $Y$ and $S$ are assumed to be symmetric and positive definite. The quadratic form $\|\hat{y}_{k+i|k} - r_{k+i}\|_Y^2 \triangleq (\hat{y}_{k+i|k} - r_{k+i})^T Y (\hat{y}_{k+i|k} - r_{k+i})$ provides a mechanism to allow different weightings on different outputs. This is similar for $\|u_{k+i|k} - u_{k+i-1|k}\|_S^2 \triangleq (u_{k+i|k} - u_{k+i-1|k})^T S (u_{k+i|k} - u_{k+i-1|k})$, which allows different penalties for different input moves. Furthermore, $u_{k|k} \triangleq u_k$ is the input applied at the current time, i.e. the input calculated during the previous period.

## 4. NUMERICAL SIMULATION

### 4.1 Simulation Setup

In this work, all simulations are completed in MATLAB/Simulink. The parameters of the quadrotor system in Eqs. (1) – (4) and (8) are listed in Table 1.

In order to compare the impacts of the suspended load mass on the quadrotor system, we discuss eleven cases with different load masses. As mentioned by most quadrotor manufacturers and designers, the quadrotor behaves inappropriately or unreasonably if the payload mass exceeds its maximum capacity. Given this, it is meaningless to evaluate our controller performance beyond the maximum payload mass. Additionally, if using the theoretical maximum load mass, the quadrotor does not have any additional power to track the desired trajectory since all its power is devoted to lifting the load. Here we define the maximum load mass that the quadrotor can carry while meeting the required trajectory requirement as 'critical motion mass ($m_{cm}$)'. To make the comparison of different controllers' performances more reasonable, in this section, we only consider the load masses not larger than critical motion mass. The performance of the system with the mass load greater than the critical motion mass will be discussed in details in the next section. According to the designed trajectory, which will be described later, the critical motion mass is 0.5 kg so that the load mass used in this section is within the range (0, 0.5kg]. All these load masses can be handled by the quadrotor. The 0kg load mass is not considered in this simulation because a 0kg load makes the cable force negligible thereby violating assumption (4). However, we can use a very light load with a value of 0.005kg to represent the 0kg load scenario. To identify the impact of the load mass on system performance, 10 different load masses are selected in a range between 0.05kg and 0.5kg evenly. That is, load masses with values of 0.005, 0.05, 0.1, 0.15, 0.2, 0.25, 0.3, 0.35, 0.4, 0.45, and 0.5 are selected for this simulation. In each case designed above, we define the quadrotor initial position and velocity as $[0,0,1.5]^T$ and $[0,0,0]^T$. The load initial position is $[0,0,0.5]^T$ and the load initial velocity is $[0,0,0]^T$, meaning the cable is straight downward, and the interconnected system is initially at equilibrium. Our control goal is to let the quadrotor track a square trajectory with a side length of 0.5m. Given the total simulation time is 75s, the quadrotor is expected to move in the X direction with 0.5m during 0-15s and then move in the Y direction with another 0.5m in 15-30s. From 30s to 45s, it moves in the –X direction with 0.5m and in next 15s it moves in the –Y direction with 0.5m. In each 15s period, the quadrotor is supposed to accelerate with $0.032\ m/s^2$ during first 2.5s, move with a constant speed of $0.08\ m/s$ for 10s, and decelerate with $0.032\ m/s^2$ for the last 2.5s. Once the quadrotor returns to the starting point, it has another 15s to stabilize. To make the expression clear and convenient, we treat the whole procedure as 5 stages. Stages 1-4 indicate four edges of the square trajectory, and stage 5 indicates the stabilization process once back to the starting point. Since the altitude z and yaw angle $\psi$ has no coupling with other states and can remain the desired value precisely, the control performance is evaluated on the quadrotor's x, y positions, the roll, pitch angles and the load's x, y positions. The control parameters are listed in Table 2.

## 4.2 Simulation Results

Several evaluation criteria are introduced to evaluate the simulation results. When these criteria appear in the following context, we will also explain them with figures to provide a better illustration.

(1) Maximum Trajectory Tracking Error $e_{max}$ (shown in Figure 5(a)): Trajectory Tracking Error indicates the difference between the desired quadrotor position and the actual quadrotor position. The maximum value among all these error values is the maximum trajectory tracking error $e_{max}$.

(2) Maximum Roll/Pitch angle $\Phi_{max}/\theta_{max}$: The maximum absolute roll/pitch angle during the total simulation. Note that $\Phi_{max}$ is equal to $\theta_{max}$ due to the symmetric desired trajectory. As a result, we will only use $\Phi_{max}$ for the evaluation later.

(3) Maximum Stabilization Time $t_{smax}$ (shown in Figure 6): The stabilization time is calculated by the difference between the first time the roll/pitch angle starts to change and the first time the roll/pitch angle entering the range [-0.2, 0.2] degrees (during the convergence process). Since there are four stabilization times in the total simulation, the maximum value among these four is defined as maximum stabilization time.

In fact, the control performances of all 11 designed cases are similar. The main differences between them exist in the three criteria above. To avoid the repeated description, we select one typical case with a 0.3kg load mass to show the control performance graphically. The differences among all 11 cases are presented using a table format. The performance figures are shown in Figure 5 – Figure 11.

Figure 5 shows the quadrotor trajectory tracking results in both X and Y directions. Because the quadrotor trajectory tracking has a similar pattern along the X direction and Y direction, we take the trajectory of the X direction in stage 1 as an exemplary illustration. In Figure 5(a), the black dotted line, red solid line, blue dashed line, and green dash-dot line represent the desired trajectory, PD, SMC, and MPC actual trajectories, respectively. During the acceleration period (0-2.5s), the quadrotor first lags behind the desired trajectory due to the suspended load's influence. After the quadrotor's desired acceleration changes to 0 (2.5s-12.5s), the quadrotor first keeps lagging (for around 0.1s) due to the movement inertia. The actual control system also needs a small time interval to respond and execute its next command. Then the quadrotor begins converging to the desired trajectory. When entering the deceleration period (12.5s-15s), the quadrotor is dragged forward by the load in front of the trajectory before returning to the desired trajectory again. As demonstrated by the final results later, the time of this process depends on the controller performance as well as the load mass. In most cases, the interconnected system is able to stabilize in stage 4 so there is no oscillation in stage 5 just like there is no oscillation in stage 1 for the quadrotor y position tracking in Figure 5(b). The load position trajectory tracking is similar. The main difference is the greater lead and lag magnitudes. This is because our control influences are only on the quadrotor so that the load's stabilization time and trajectory tracking are asymptotical.

The acceleration profile is shown in Figure 6. The orange solid line represents the desired acceleration, and the light blue solid line indicates the actual acceleration. Note that all our controllers are based on position errors (i.e. position control), so the acceleration changes gradually rather than instantly. The acceleration increases at the desired acceleration period. However, it has oscillation due to the load's influence. When the desired acceleration disappears during 2.5s-12.5s, the quadrotor still needs to accelerate to compensate for the trajectory tracking error. The oscillation in this period is also the stabilization process of the interconnected system. The deceleration part is similar to the acceleration part, except that the stabilization process is longer than the acceleration part. This is because the load has a non-zero velocity when the interconnected system tries to stabilize from 12.5s. As a result, the system needs more efforts to eliminate this energy, which leads to a longer time frame.

To further quantify the trajectory tracking performance, Figure 7 illustrates the quadrotor trajectory tracking errors between the desired and actual position in both the X and Y directions. It shows the errors of PD, SMC and MPC controllers in the order from top to bottom, respectively. The very bottom figure shows the desired acceleration profile to make each period and stage more readable. Take, for instance, the error in the X position. The error first increases in the acceleration phase and keeps increasing to the maximum trajectory tracking error $e_{max}$ about 0.1s after the acceleration phase. Then the error begins to decrease and converge to zero. This period can be seen as the first error oscillation period. When the quadrotor enters the deceleration part, the error starts another oscillation period. The maximum error in this period is smaller than the first one since the quadrotor is not static at the beginning of this oscillation period thereby making the adjusting process easier, but with a longer stabilization time. From this figure, we can see MPC has the smallest maximum error, 0.0171m, while SMC has an error of 0.0396m, which is smaller than PD's 0.0506m. In addition, the convergence of MPC during the constant speed period is fastest. While SMC is also able to converge with a longer time, the PD controller still has some small oscillations even near the end of this period.

Figure 8 shows the quadrotor roll and pitch angles. All angles of the three methods are confirmed to be inside the range [-4, 4] degrees. Small angle approximation works well for all angle values between [-6, 6] degrees. Therefore, the small roll and pitch angles ensure no violations of the low-speed and small-angle movement assumptions. As seen in Eqs. (13) and (14), roll is related to the Y movement and pitch is related to the X movement. This is why the roll angle changes when the quadrotor has acceleration in the Y direction. $\Phi_{max}/\theta_{max}$ appears at the beginning of the constant movement period since that is when the acceleration is maximum. It matches the acceleration profile in Figure 6. Figure 8 also shows the PD controller has the smallest $\Phi_{max}/\theta_{max}$ while MPC has the largest. The possible reason for this is that MPC and SMC sacrifices the angle overshoot to the stabilization time. However, considering all angles are still close and inside the small range, we should focus more on the stabilization time, which reflects the dynamic adjusting process more significantly. As explained at the beginning of section 4.2, there are four stabilization times. In Figure 8(a), for example, they begin from 15s, 27.5s, 45s, and 57.5s, respectively. In fact, the stabilization time starting at 15s is equal to the one starting at 45s due to a symmetric trajectory design. They are shorter than the ones starting at 27.5s and 57.5s, which are also equal. Thus, the stabilization time starting at 27.5s and 57.5s can both be called the maximum stabilization time $t_{smax}$. For expression convenience, however, we only use the one starting at 27.5s to indicate $t_{smax}$ in this paper. The increase of $t_{smax}$ in the order of PD, SMC and MPC from Figure 8 verifies MPC has the best performance as well.

In order to have a better view of the control performance, the PD results are drawn in 3D space for an instance as shown in Figure 9. The black dashed lines are desired quadrotor and load trajectories, the orange dotted line with squares is actual quadrotor path, and the dark red dotted line with triangles is actual load path. The light blue cross and circle symbol represent the quadrotor and the load, respectively. According to the zoom-in part in this figure, we can see there is an angle between the actual and desired cable position. The reasons causing this angle are the load has a larger oscillation than the quadrotor and the relative positions r and s (modeled in Eq. (8) and Figure 2) are non-zero during the dynamic adjusting process as shown in Figure 10. The relative position r and s values describe the dynamic adjusting process. The maximum relative position values indicate the load's maximum oscillation angles. The time their values converge to zero represents how long the load can go back to equilibrium status. From Figure 10, we can see MPC has the minimum relative position r and s values while PD has the largest. This means the load oscillation range of PD is larger than both SMC and MPC so that the maximum oscillation angle exists in the PD control results. Moreover, MPC has the shortest converging time at each period during 0-75s mentioned in the trajectory design description in section 4.1. On the contrary, PD needs a longer time to reach the equilibrium.

Three tracking paths on X-Y plane are shown in Figure 11. The black dashed line, red dotted line, blue solid line, and green dash-dot line are the desired trajectory, and actual trajectories of PD, SMC, and MPC, respectively. From this figure, we can see both PD and SMC exceed the desired corners when changing the quadrotor movement direction. However, the SMC's exceeding part is smaller than PD. The oscillation of SMC is also smaller than PD after the direction changes. In contrast, MPC tracks those corners quite well and has nearly negligible oscillation. This result also coincides with the error change around 15s in Figure 7. The load path result is similar, but it is worse than the quadrotor path. This is because we focus on the quadrotor control instead of the load. Figure 11 again confirms the MPC has the best result while the PD control has the worst one.

The above is a typical exemplary case to illustrate three control results graphically. As mentioned in Section 4.1, we discussed eleven cases starting from a 0.005kg load and then with a 10% incremental increase to 0.05kg. Table 3 and Table 4 present more simulation data to help analyze the effects of different load masses.

In each row of Table 3, MPC has the smallest maximum trajectory tracking errors for the same load mass. SMC is about 0.022m more than MPC and about 0.011m less than the PD controller. In other words, SMC reduces the tracking error by about 21.9% more than PD while MPC reduces the tracking error by about 66.4% more than PD. This also means the MPC's deviation from the desired trajectory is smallest so the MPC has the best tracking performance. The reason causing the better performances of SMC and MPC is their own feature of better robustness against external disturbances [13, 32]. Furthermore, the Kalman Estimator inside the MPC design helps eliminating the disturbance more actively so that MPC's performance is better than SMC.

However, the maximum trajectory tracking error in Table 3 remains very close for each controller as the load mass increases. This insignificant change indicates the load mass uncertainty's major influence is not in the trajectory tracking error $e_{max}$ as long as the load mass is not larger than the critical motion mass. Instead, it affects dynamic responses, including the maximum stabilization time $t_{smax}$ and maximum angle $\Phi_{max}$, as shown in Table 4. $\Phi_{max}$ and $t_{smax}$ increase as the load mass increases and the change of $t_{smax}$ is especially significant. To be specific, the maximum $t_{smax}$ increments for PD and SMC are 8.51s and 5.65s, respectively while MPC has the same $t_{smax}$. In other words, SMC and MPC reduce the stabilization time by about 42.9% and 72.1% more than PD. As for $\Phi_{max}$, however, MPC has the largest value while SMC has the smallest. A possible reason for this is that MPC sacrifices its angle adjusting for the stabilization time adjusting. But the maximum $\Phi_{max}$ increment of MPC is 0.21 degrees, which is very close to SMC's 0.10 degrees and much larger than PD's 1.71 degrees. Also considering the significant effect in reducing the stabilization time, MPC handles the load mass uncertainty best while SMC is better than PD in handling this uncertainty.

Combining Table 3 and Table 4, SMC and MPC have not only a favorable tracking performance but also a good ability to deal with the load uncertainty. This means such robust controller design can be very useful in the practical uncertain load transportation if users want the performance to be stable. Especially for the varying load situation, it is not convenient or even impossible to change control parameters in the real-time flying, this stable performance can greatly help improving the quadrotor safety. If there is a very strict requirement on the quadrotor's trajectory, a more robust controller can provide better performance under uncertain loads.

**5. FURTHER DISCUSSION**

Our studies show that a critical motion mass ($m_{cm}$) exists for the quadrotor while maintaining the expected performance. In fact, this critical mass limit exists when the interconnected system moves together with the same acceleration. If analyzing all the forces on the quadrotor, there are a total of three: the total thrust force, the force caused by the quadrotor's gravity, and the force caused by the cable (i.e. caused by the load's gravity and inertia). The resultant force from these three forces leads to an acceleration for the quadrotor and load. Based on these force analyses, the critical motion mass $m_{cm} = (U_1 \cos(\text{atan}(a_{desired}/g)))/g - m_q$, where $U_1$ is the total thrust in Eq. (6), $m_q$ is the quadrotor mass, and $a_{desired}$ is the quadrotor's desired acceleration.

In addition, this critical motion mass divides the load impacts into two parts. If the actual load mass $m_L \leq m_{cm}$, the quadrotor can provide enough forces to handle both the load mass and expected motion. In this case, the mass change mainly results in the stabilization as discussed before. If $m_L > m_{cm}$, the quadrotor cannot meet the desired acceleration anymore since the quadrotor needs to devote most or even all its force to lift the load first. Instead, the maximum acceleration the quadrotor can reach now is $a_{max} = \sqrt{U_1^2 - \left((m_q + m_l)g\right)^2}\big/(m_q + m_l)$. This can also be regarded as a critical mass-related acceleration ($a_{cm}$). Given a fixed load mass, $a_{cm}$ should not be less than the quadrotor's desired acceleration $a_{desired}$ to maintain an expected performance. If $a_{cm} < a_{desired}$, there is a significant lag between the expected and actual trajectory. This lag keeps the tracking error increasing as long as $a_{cm} < a_{desired}$.

For a better illustration of what occurs when the load mass is in the situation discussed above, Figure 12 shows the trajectory differences between a 0.5kg and 0.55kg load mass. For the desired trajectory in this figure, we only use stage 1 of the X direction movement in previous simulations. This means the quadrotor is still supposed to move with 1m within 15s along the X direction. After reaching the destination, the quadrotor remains at that position for the rest of the time (15s-75s). This experiment design does not affect the result since other stages are symmetric. Also note the acceleration for this desired trajectory corresponds to $m_{cm} = 0.5kg$, so the quadrotor should not be able to reach this acceleration when carrying a 0.55kg load.

In Figure 12, the red dotted line indicates the desired quadrotor trajectory. The blue dash-dot line and green dashed line represent the actual quadrotor trajectory with a 0.5kg and 0.55kg load, respectively. While the quadrotor with a 0.5kg load can track the desired path well, there is a significant lag for the quadrotor with a 0.55kg load. The tracking error keeps increasing till 15s when the quadrotor is supposed to stop moving. After 15s, however, the quadrotor needs to keep moving in reality since the real position does not reach the expected destination. In fact, the quadrotor reaches the final destination around 19s, which means it uses additional 4 seconds to make up the lagged distance. An adjusting process still exists when the quadrotor with a 0.55kg load reaches the desired position. But this process indicated by the oscillation is smoother than the 0.5kg load case. This is because the quadrotor has a lower speed around the destination so that the interconnected system needs less effort and time to remove the speed's effect. In sum, the quadrotor can still reach the final destination if given enough time. But if considering a limit time, the quadrotor cannot track the desired trajectory or reach the destination anymore.

## 6. CONCLUSIONS

In this paper, we have investigated the influence of uncertain load mass on the quadrotor. With the performance comparison among three different controllers, we find SMC and MPC reduce the stabilization time by about 42.9% and 72.1% and reduce the maximum trajectory tracking error by 21.9% and 66.4% when compared with the PD controller. These results of SMC and MPC verify the importance of the controller's robustness against disturbances. Therefore, a controller with a strong robustness against disturbances is suggested for the practical uncertain load transportation. The results also indicate the main impact of the load mass uncertainty is on the quadrotor's stability rather than the trajectory tracking error. If there is a strict requirement on the load transportation's stability or fast response, a robust/fast controller design like MPC is suggested. We have also analyzed the consequences when the quadrotor cannot handle the load mass and the tracking requirement simultaneously. If the quadrotor's desired acceleration is larger, then the quadrotor's critical motion mass is smaller.

In the next stage, we want to verify the simulation results in the real quadrotor systems. Before the real experiment, it is meaningful to reduce the computation cost of MPC. The current MPC code executing time is about 4-5 times of PD and SMC. Successfully reducing the computation cost will provide MPC a more promising real-time implementation with large prediction horizon and control horizon. Additionally, it seems an estimator inside a controller design may play an important role in improving the performance. The investigation of the estimator's influence could be used for further work in this area. We are also interested in optimizing the load trajectory more than the quadrotor trajectory since some practical applications only require the load position accuracy while having a big movement relaxation for the quadrotor.

**NOMENCLATURE**

| | |
|---|---|
| $\boldsymbol{p}$ | quadrotor position |
| $\boldsymbol{v}$ | quadrotor linear translational velocity |
| $m_q$ | quadrotor mass |
| $m_L$ | load mass |
| $g$ | acceleration of gravity |
| $\boldsymbol{R}$ | rotation matrix from frame {W} to frame {I} |
| $f$ | upwards thrust |
| $\boldsymbol{\Omega}$ | angular velocity |
| $\boldsymbol{S}(\cdot)$ | skew-symmetric operator |
| $\boldsymbol{I}$ | body-fixed inertia matrix |
| $\boldsymbol{T}$ | applied moments |
| $F_i$ | thrust force generated by rotor i |
| $Q_i$ | moment generated by rotor i |
| $U_i$ | quadrotor system input |
| $r$ | load x position |
| $s$ | load y position |
| $\zeta$ | load z position |
| $L$ | cable length |
| $F_{Ci}$ | disturbance caused by load to quadrotor in i (x, y, z) direction |
| $F_d$ | disturbance for a general discrete-time system |
| $S_i$ | sliding surfaces |
| $\lambda_i$ | sliding control parameter |
| $k_i$ | sliding control parameter |
| DOF | degree of freedom |
| *PD* | proportional-derivative |
| SMC | sliding mode control |
| MPC | model predictive control |

## Figure Captions List

Fig. 1        Schematic of quadrotor dynamics

Fig. 2        Schematic of quadrotor dynamics with a slung load

Fig. 3        Overall Control Scheme. According to the expected and current position, the SMC/MPC position controller determines the desired roll and pitch angles for the SMC/MPC Attitude controller to track. Based on position and attitude controllers' control signals, the quadrotor dynamics block determines the new states, which are also the inputs of the load dynamics and the position controller for another round of control.

Fig. 4        Block diagram of MPC

Fig. 5        Position comparison

Fig. 6        Acceleration profile

Fig. 7        Position error comparison

Fig. 8        Quadrotor roll and pitch angles

Fig. 9        3D vision of the trajectory. The black dashed lines are desired quadrotor and load trajectories while the orange dotted line with squares is actual quadrotor path and the dark red dotted line with triangles is actual load path. The light blue cross and circle symbol represent the quadrotor and the load respectively.

Fig. 10       Relative position comparison

## Table Caption List

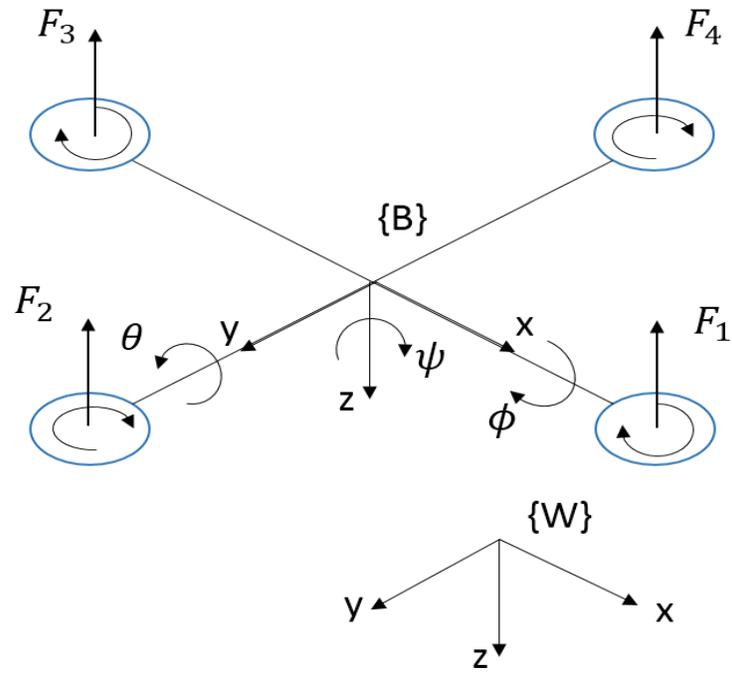

Figure 1: Schematic of Quadrotor Dynamics

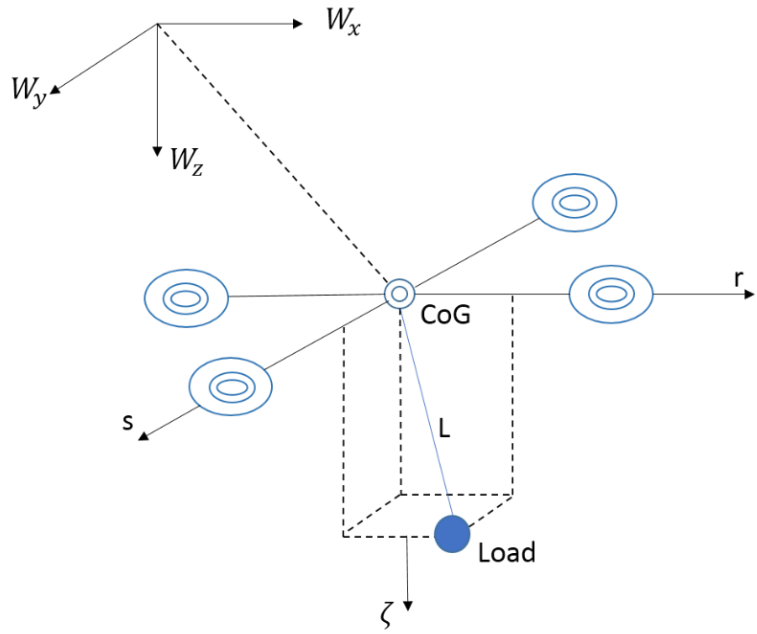

Figure 2: Schematic of quadrotor dynamics with a slung load

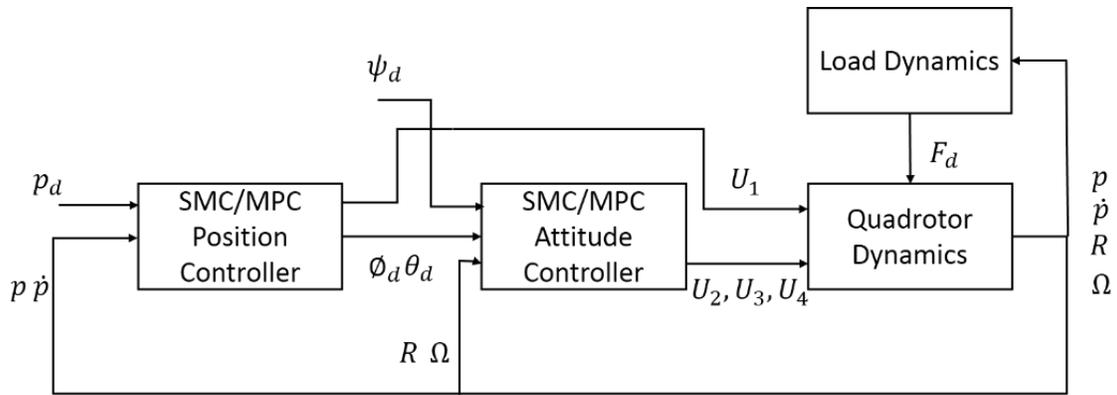

Figure 3: Overall Control Scheme. According to the expected and current position, the SMC/MPC position controller determines the desired roll and pitch angles for the SMC/MPC Attitude controller to track. Based on position and attitude controllers' control signal, the quadrotor dynamics block determines the new states, which are also the inputs of the load dynamics and the position controller for another round of control.

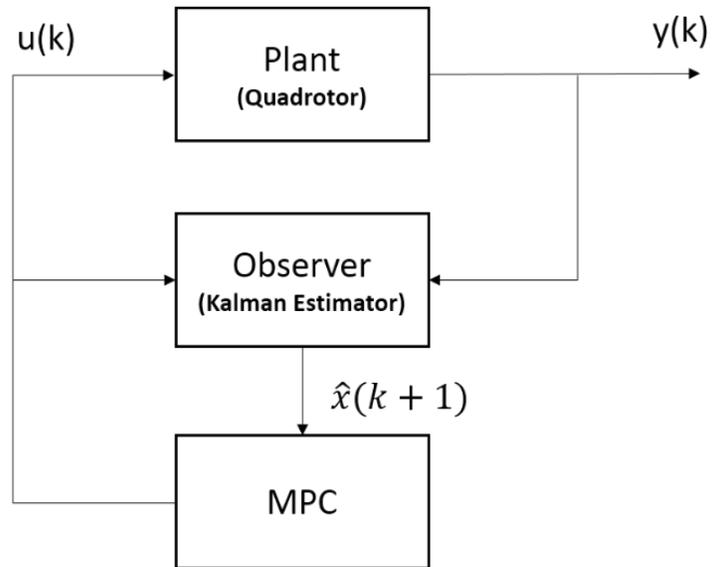

Figure 4: Block Diagram of MPC

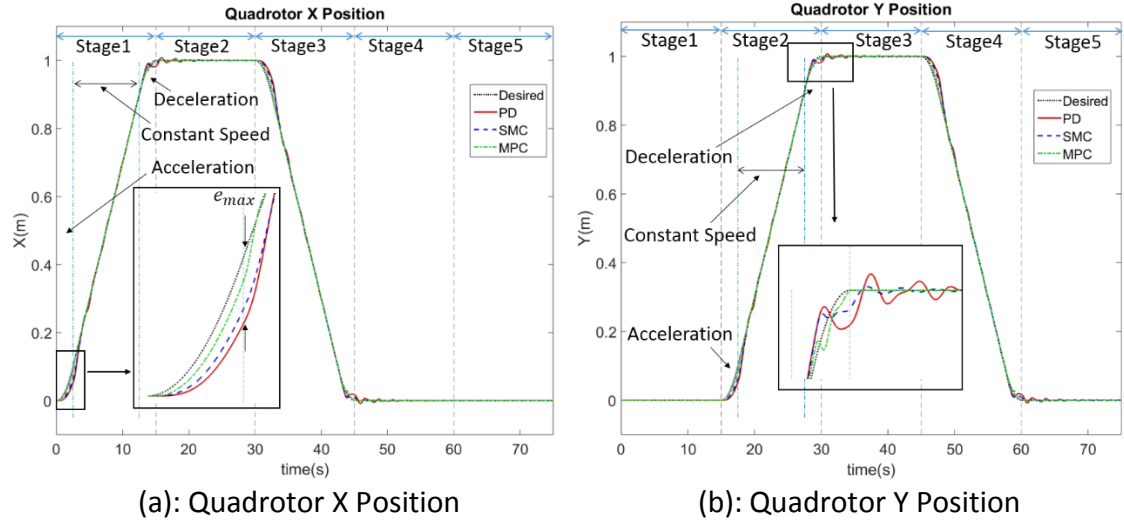

(a): Quadrotor X Position       (b): Quadrotor Y Position

Figure 5: Position Comparison

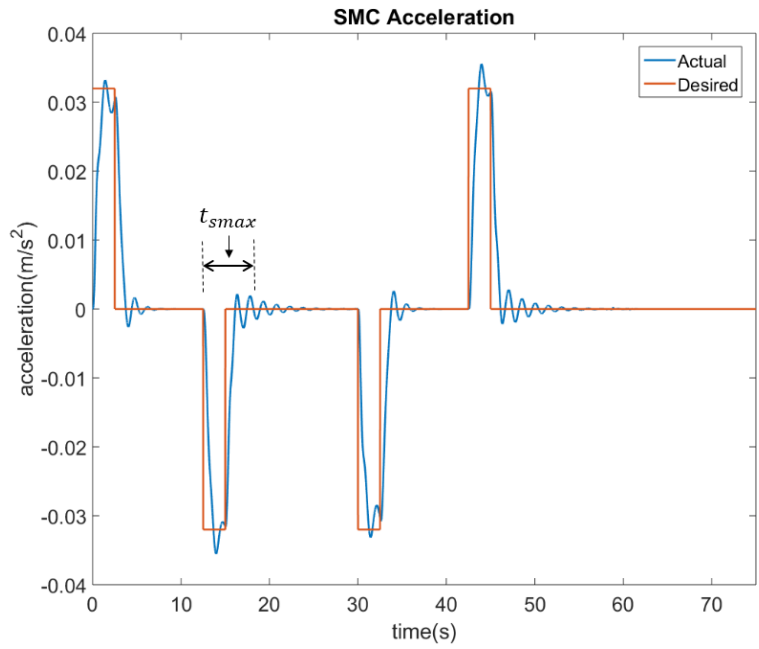
Figure 6: Acceleration Profile

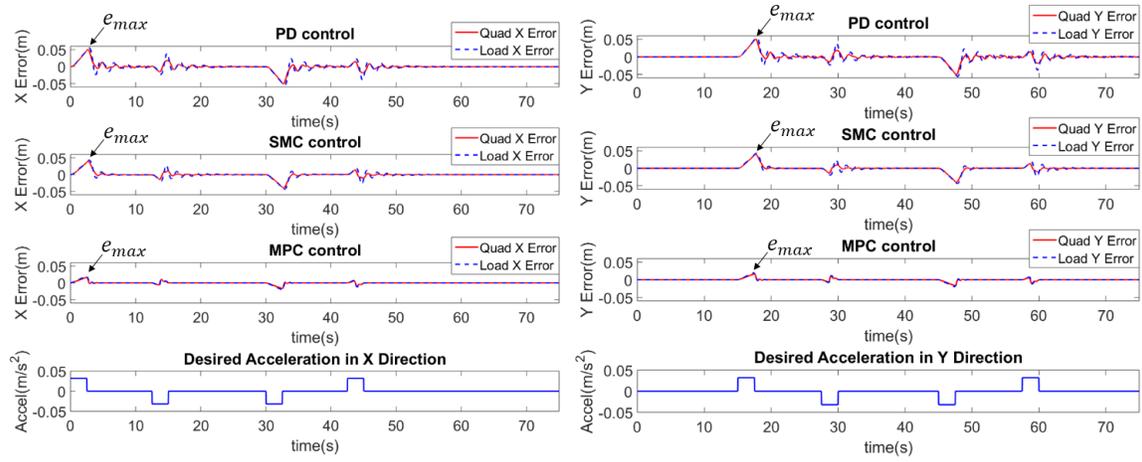

(a): Position X Error    (b): Position Y Error

Figure 7: Position Error Comparison

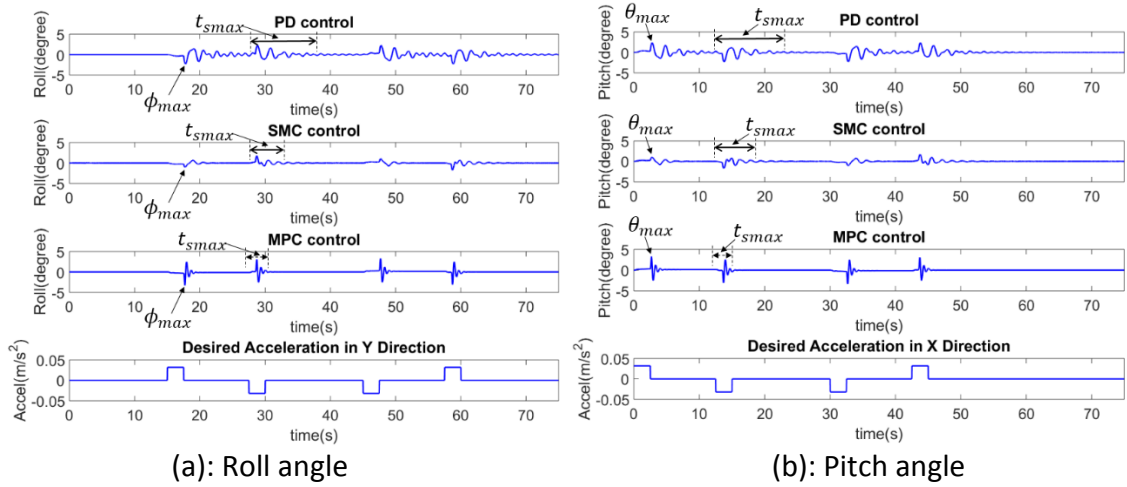

(a): Roll angle  (b): Pitch angle

Figure 8: Quadrotor Roll and Pitch angles

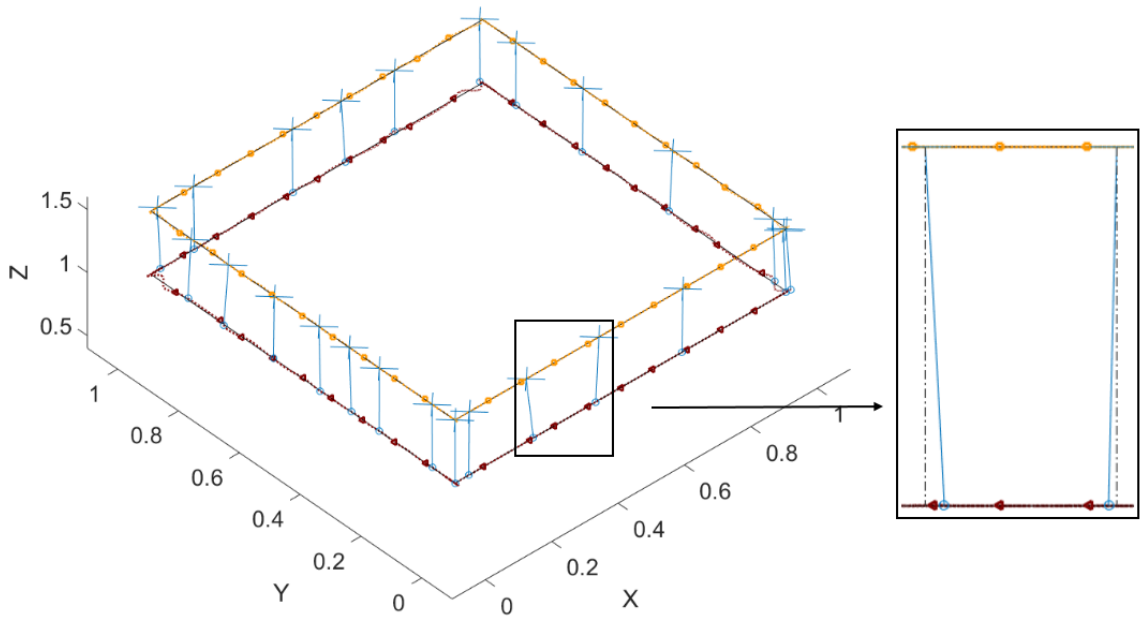

Figure 9: 3D vision of the trajectory. The black dashed lines are desired quadrotor and load trajectories while the orange dotted line with squares is actual quadrotor path and the dark red dotted line with triangles is actual load path. The light blue cross and circle symbol represent the quadrotor and the load respectively.

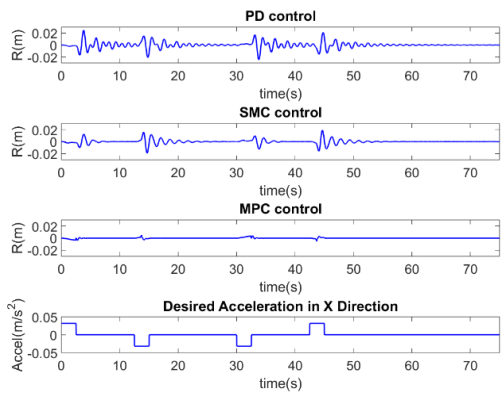 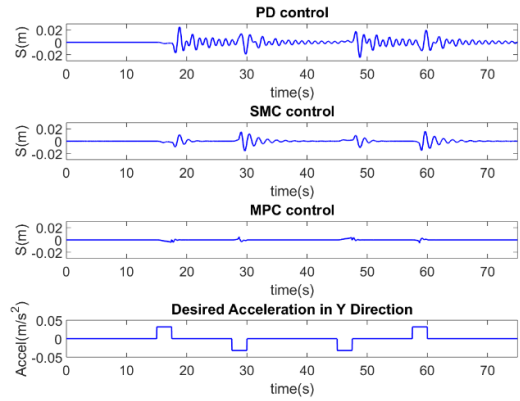

(a): Relative Position r          (b): Relative Position s

Figure 10: Relative Position Comparison

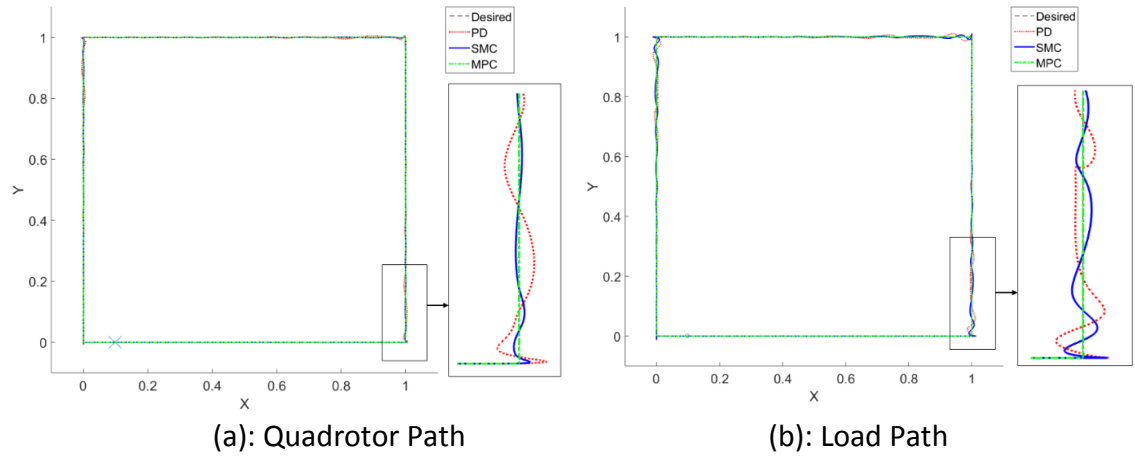

(a): Quadrotor Path          (b): Load Path

Figure 11: Quadrotor and Load Actual Path Following the Square Trajectory

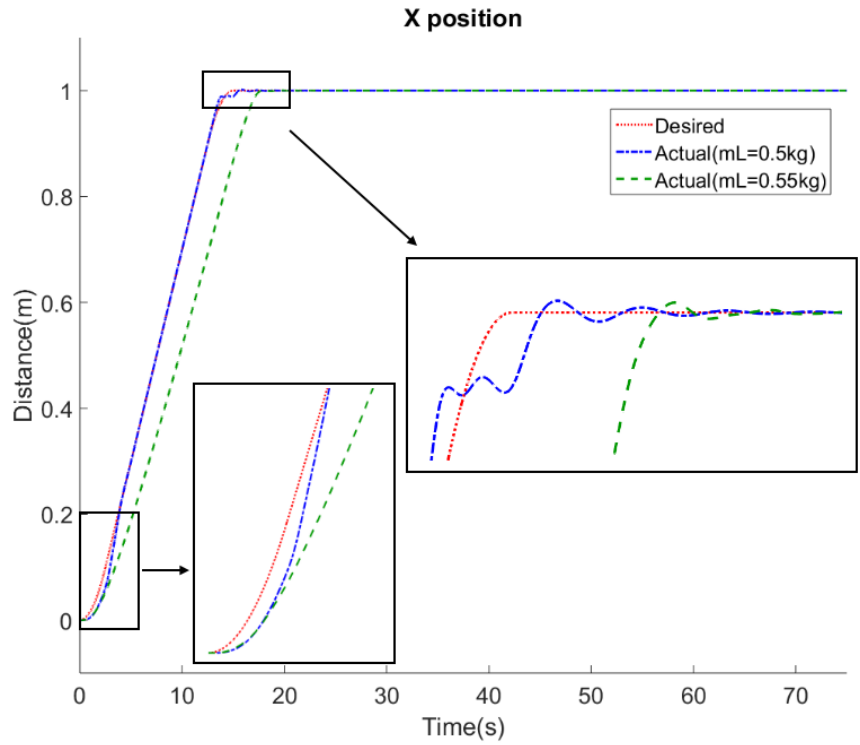

Figure 12: Performance Comparison Between 0.5kg and 0.55kg Load Mass

Table 1: Quadrotor Simulation Parameters

| Parameter | Value | Description |
|---|---|---|
| $I_x, I_y$ | 7.5*10^(-3) $kg/m^2$ | Quadrotor moment of inertia around X axis and Y axis |
| $I_z$ | 1.3*10^(-2) $kg/m^2$ | Quadrotor moment of inertia around Z axis |
| $m_q$ | 1kg | Quadrotor mass |
| b | 3.13*10^(-5) | Thrust factor |
| d | 7.5*10^(-7) | Drag factor |
| $l$ | 0.25m | Distance between the rotor center and quad center |
| L | 0.5m | The cable length |
| $M_{max}$ | 0.6kg | Maximum payload mass |

Table 2: Control Parameters

| Parameter | Value | Description |
|---|---|---|
| $K_{px}, K_{py}$ | 10 | Proportional gain for x and y |
| $K_{pz}$ | 20 | Proportional gain for z |
| $K_{dx}, K_{dy}$ | 8 | Derivative gain for x and y |
| $K_{dz}$ | 15 | Derivative gain for z |
| $K_{pp}, K_{pt}$ | 50 | Proportional gain for $\phi$ and $\theta$ |
| $K_{pps}$ | 20 | Proportional gain for $\psi$ |
| $K_{dp}, K_{dt}$ | 20 | Derivative gain for $\phi$ and $\theta$ |
| $K_{dps}$ | 15 | Derivative gain for $\psi$ |
| $k_i$ | [0.4, 0.4, 0.4, 0.6, 0.6, 0.4] | $k$ related parameter in the order $\phi, \theta, \psi$, x, y and z in Eq. (12) |
| $\lambda_i$ | [0.5, 0.5, 0.5, 2.25, 2.25, 5] | $\lambda$ related parameter in the order $\phi, \theta, \psi$, x, y and z in Eq. (12) |
| N | 25 | MPC prediction horizon |
| M | 25 | MPC control horizon |

Table 3: Maximum Tracking Error for Different Load Masses

| Load Mass (kg) | PD Maximum Trajectory Tracking Error $e_{max}$ (m) | SMC Maximum Trajectory Tracking Error $e_{max}$ (m) | MPC Maximum Trajectory Tracking Error $e_{max}$ (m) |
|---|---|---|---|
| **0.005** | 0.0500 | 0.0387 | 0.0163 |
| **0.05** | 0.0501 | 0.0389 | 0.0164 |
| **0.1** | 0.0502 | 0.0390 | 0.0166 |
| **0.15** | 0.0503 | 0.0392 | 0.0167 |
| **0.2** | 0.0504 | 0.0393 | 0.0168 |
| **0.25** | 0.0505 | 0.0395 | 0.0170 |
| **0.3** | 0.0506 | 0.0396 | 0.0171 |
| **0.35** | 0.0507 | 0.0397 | 0.0173 |
| **0.4** | 0.0508 | 0.0398 | 0.0174 |
| **0.45** | 0.0509 | 0.0399 | 0.0175 |
| **0.5** | 0.0510 | 0.0401 | 0.0177 |

Table 4: Maximum Angle and Stabilization Time for Different Load Masses

| Load Mass (kg) | PD Maximum Angle $\phi_{max}$ (degree) | PD Maximum Stabilization Time $t_{smax}$ (s) | SMC Maximum Angle $\phi_{max}$ (degree) | SMC Maximum Stabilization Time $t_{smax}$ (s) | MPC Maximum Angle $\phi_{max}$ (degree) | MPC Maximum Stabilization Time $t_{smax}$ (s) |
|---|---|---|---|---|---|---|
| 0.005 | 1.58 | 5.25 | 1.63 | 2.51 | 3.51 | 2.73 |
| 0.05 | 1.75 | 8.52 | 1.65 | 4.91 | 3.52 | 2.73 |
| 0.1 | 1.78 | 9.08 | 1.66 | 5.53 | 3.55 | 2.73 |
| 0.15 | 1.95 | 9.73 | 1.66 | 5.71 | 3.57 | 2.73 |
| 0.2 | 2.06 | 9.88 | 1.67 | 5.77 | 3.59 | 2.73 |
| 0.25 | 2.16 | 10.52 | 1.68 | 6.22 | 3.61 | 2.73 |
| 0.3 | 2.32 | 10.96 | 1.69 | 6.24 | 3.63 | 2.73 |
| 0.35 | 2.73 | 11.07 | 1.70 | 6.62 | 3.65 | 2.73 |
| 0.4 | 3.00 | 12.86 | 1.71 | 6.89 | 3.67 | 2.73 |
| 0.45 | 3.02 | 13.57 | 1.71 | 7.53 | 3.69 | 2.73 |
| 0.5 | 3.29 | 13.76 | 1.73 | 8.16 | 3.72 | 2.73 |